\documentclass[sigconf,authorversion,nonacm]{acmart}

\usepackage{array}
\usepackage{caption}
\usepackage{courier}
\usepackage{diagbox}
\usepackage{graphicx}
\usepackage{helvet}
\usepackage{multirow}
\usepackage{natbib}
\usepackage{subcaption}
\usepackage{times}
\begin{document}

\title{Clustering augmented Self-Supervised Learning: An application to Land Cover Mapping}

\author{Rahul Ghosh}
\email{ghosh128@umn.edu}
\affiliation{%
  \institution{University of Minnesota}
  \city{Minneapolis}
  \state{MN}
  \country{USA}
}

\author{Xiaowei Jia}
\email{xiaowei@pitt.edu}
\affiliation{%
  \institution{University of Pittsburgh}
  \city{Pittsburgh}
  \state{PA}
  \country{USA}
}

\author{Chenxi Lin}
\email{lin00370@umn.edu}
\affiliation{%
  \institution{University of Minnesota}
  \city{Minneapolis}
  \state{MN}
  \country{USA}
}

\author{Zhenong Jin}
\email{jinzn@umn.edu}
\affiliation{%
  \institution{University of Minnesota}
  \city{Minneapolis}
  \state{MN}
  \country{USA}
}

\author{Vipin Kumar}
\email{kumar001@umn.edu}
\affiliation{%
  \institution{University of Minnesota}
  \city{Minneapolis}
  \state{MN}
  \country{USA}
}

\begin{abstract}
Collecting large annotated datasets in Remote Sensing is often expensive and thus can become a major obstacle for training advanced machine learning models. Common techniques of addressing this issue, based on the underlying idea of pre-training the Deep Neural Networks (DNN) on freely available large datasets, cannot be used for Remote Sensing due to the unavailability of such large-scale labeled datasets and the heterogeneity of data sources caused by the varying spatial and spectral resolution of different sensors. Self-supervised learning is an alternative approach that learns feature representation from unlabeled images without using any human annotations. In this paper, we introduce a new method for land cover mapping by using a clustering based pretext task for self-supervised learning. We demonstrate the effectiveness of the method on two societally relevant applications from the aspect of segmentation performance, discriminative feature representation learning and the underlying cluster structure. We also show the effectiveness of the  active sampling using the clusters obtained from  our method in improving the mapping accuracy given a limited budget of annotating.
\end{abstract}

\maketitle

\section{Introduction}
\label{Sec:Introduction}
Global demand for land resources to support human livelihoods and well-being through food, fiber, energy and living space will continue to grow in response to the population expansion and socioeconomic development. This poses a great challenge to the human society, given the increasing competition for land from the need to maintain other essential ecosystem services. Addressing this challenge will require timely information on land use and land cover changes, e.g., the conversion of forest to farmland or plantations, the loss of productive cropland due to urbanization, and the degradation of soil due to inappropriate management practices.

\begin{figure}[!h]
    \centering
    \includegraphics[width=\columnwidth]{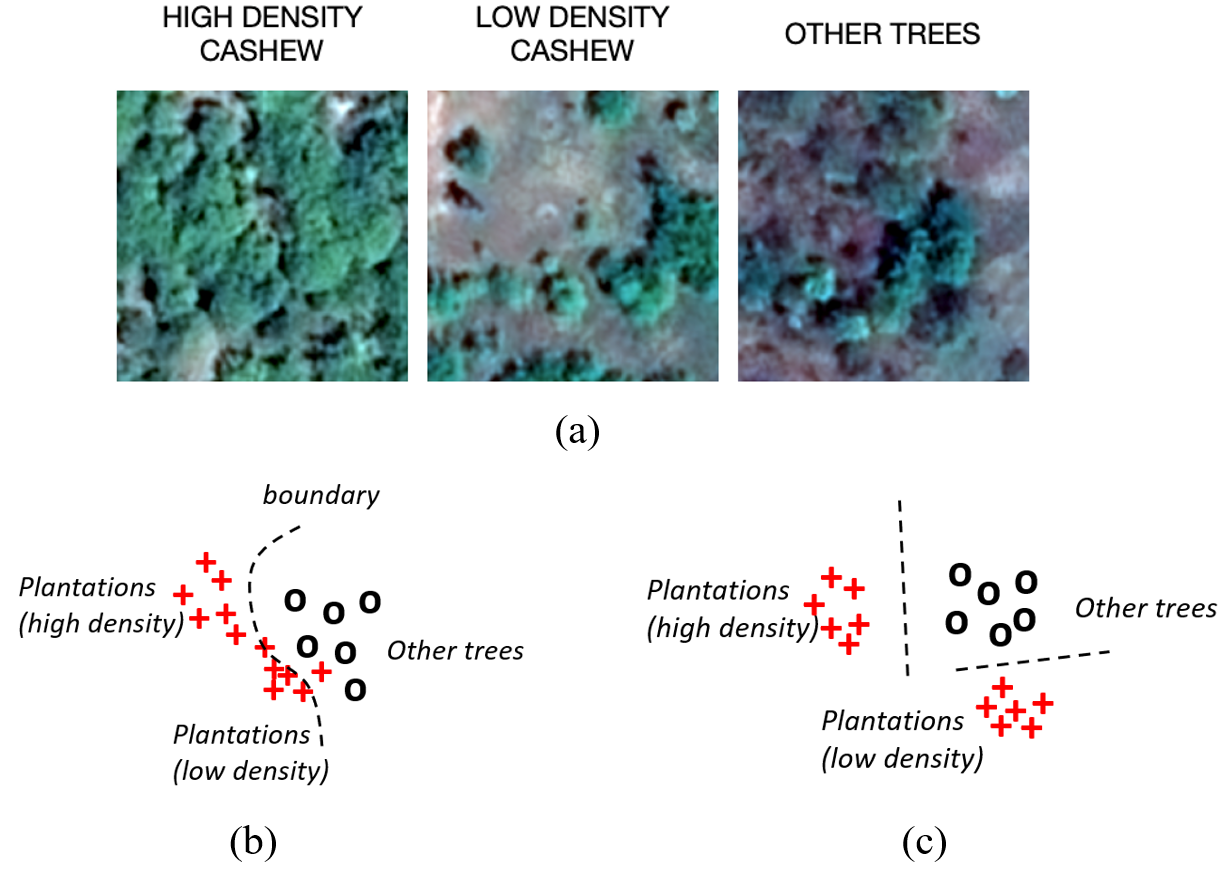}
    \caption{(a) Examples of high density Cashew plantations, low density Cashew plantations and other trees. We also show the decision boundaries  (b) learned by traditional methods  and (c) after the clustering structure is informed.}
    \label{fig:Examples}
\end{figure}

Recent advances in storing and processing remote sensing data collected by sensors onboard aircrafts or satellites provide tremendous potential for mapping a variety of land covers, including plantations~\cite{jia2017incremental}, agricultural facilities~\cite{handan2019deep}, roads~\cite{xie2015transfer}, buildings~\cite{nayak2020semi},  and many more \cite{karpatne2016monitoring}. Accurate mapping of these land covers can provide critical information at desired spatial and temporal scales to assist in decision making for development investment and sustainable resource management.

Given the success of machine learning, especially deep learning, in the domain of computer vision (e.g., image segmentation), researchers have found a lot of promise for using these techniques in automated land cover mapping at large scale through analysis of remote sensing data. Existing works have mostly focused on  the supervised learning setup which requires ample labeled data. However, collecting land cover labels is often expensive and requires expert staff, equipment, and in-field measurements and thus can become a major obstacle for training advanced machine learning models.

One common approach to deal with limited availability of labeled datasets is to pre-train an ML model on existing large labels data sets for a related problem, and then refine it using a small number of labeled samples for the problem of interest. For example, models for image recognition are first trained using large-scale datasets like ImageNet \cite{deng2009imagenet} and then are fine-tuned on the limited-size dataset for the downstream task~\cite{huh2016makes}. However, such approaches cannot be used for remote sensing due to the difference in the spectral bands captured by different satellites and such large-scale labeled datasets for capturing all the data modalities are either unavailable or these efforts are still in nascent stage, resulting in the need for more research.

Self-supervised learning is an alternative approach that learns feature representation from unlabeled images. Numerous methods have been proposed under this paradigm where the central idea is to propose various pretext tasks for the network to solve, in the hope that the network will learn important feature representations by minimizing the objective function of the pretext task, such as inpainting patches \cite{pathak2016context} and image colorization \cite{zhang2016colorful,larsson2017colorization}. The representation learned by these techniques can be transferred to a classification/segmentation model.

However, existing self-supervised learning methods can be less helpful for remote sensing data since the pretext tasks they create, e.g., colorization~\cite{vincenzi2020color}, do not make full use of all the spectral bands of remote sensing data to  capture the land cover heterogeneity. For example, the identification of cashew plantations (Fig.~\ref{fig:Examples} (a)) requires differentiating other trees from all types of cashew plantations with varying density.  High-density plantations are easily separable with other trees while low-density plantations are more likely to be confused with other trees. These self-supervised learning methods can learn similar representation between low-density plantations and other trees, which can cause potential confusion amongst classes. This poses a challenge for the segmentation model to learn a decision boundary that can correctly classify all the modes in each class during the fine-tuning process (Fig.~\ref{fig:Examples} (b)). Intuitively, if we can detect these modes by leveraging the information from all the spectral bands and inform the segmentation model of the obtained clustering structure, the segmentation model can easily learn decision boundaries to separate different classes as long as we have a few representative samples from each mode (Fig.~\ref{fig:Examples} (c)).

In this paper, we develop a self-supervised learning framework, Clustering-Augmented Segmentation (CAS), which uses clustering to capture underlying land cover heterogeneity. In particular, our clustering algorithm is inspired by DEC~\cite{xie2016unsupervised}, which is a representation learning method for image classification. Although optimizing the clustering at image-patch level  improves the classification, it results in the loss of the fine-level details which severely degrades segmentation performance. To address this issue, we build an auto-encoder-based framework which promotes the discriminative representation learning by optimizing the clustering structure over image patches while also preserving the local pixel-wise information for reconstruction. Here the clustering structure helps better represent heterogeneous land covers while the pixel-wise information is essential for improving the segmentation accuracy. We define a loss function that combines the image patch-level clustering loss and the pixel-level reconstruction loss and then iteratively refine the obtained clustering and learning representations. It is  noteworthy that our proposed method can also incorporate other clustering methods to capture land cover heterogeneity. 

We show the superiority of our method over existing self-supervised learning methods in two societally relevant applications, cashew plantation mapping and crop detection. We have demonstrated the effectiveness of the proposed method in learning both discriminative feature representation and the underlying clustering structure. We also conduct active sampling to show the potential of achieving high mapping accuracy given a limited budget of annotating.

Our contributions can be summarized as follows:
\begin{itemize}
    \item We develop a self-supervised learning framework that leverages DEC to capture land cover heterogeneity.
    \item We have demonstrated the effectiveness of the proposed method in learning with small labeled data in the context of two applications of great societal relevance.
    \item We release the code and dataset used in this work to promote reproducibility~\footnote{https://drive.google.com/drive/folders/\\1Faf7m4eO7y30g9CeyHqlelGaJwms7y9A?usp=sharing}.
\end{itemize}

\section{Related Work}
\label{Sec:Related Work}
\subsection{Land Use and Land Cover mapping}
Mapping land use and land cover (LULC) changes is essential for managing natural resources and monitoring the impact of changing climate. Recent works \cite{LULCsurvey} have explored deep learning techniques like feed forward neural networks (FFNN)~\cite{zhou2008use}, CNN~\cite{hu2018deep,stoian2019land}, LSTM~\cite{jia2017incremental} for LULC mapping. CNNs have been shown to be effective in  extracting both spectral and spatial information, whereas RNN and LSTM make use of the temporal information in modeling land cover transitions and have shown promising performance in sequence labelling. Land cover mapping can also be framed as a semantic segmentation problem \cite{ulmas2020segmentation,su2019land,saralioglu2020semantic}, where each pixel in an aerial/satellite image is classified as a land cover class. One of the most widely models in semantic segmentation is Fully Convolutional Network (FCN)~\cite{long2015fully}, which supplements the output of the deeper layers with that of the shallower layers to increase the resolution of the prediction. Based on this idea, several modifications to FCN were proposed in recent years such as SegNet \cite{badrinarayanan2017segnet}, DeconvNet \cite{noh2015learning} and UNet \cite{ronneberger2015u}. In this work, we adopt the UNet architecture, which consists of two paths, contraction path (encoder) and symmetric expanding path (decoder). The encoder consists of a stacked set of convolutional and max-pooling layers, that captures the context and a semantic understanding of the image. The decoder involves convolutional and upconvolutional layers to generate precise label maps from the output of the encoder. 

LULC mapping differs from the standard semantic segmentation  in several ways. First, due to the heterogeneity in the land covers, the same class can look different in different areas and thus each class can have multiple  modes/subclasses. Many of these land cover classes/subclasses cannot be easily distinguished using only RGB channels but require information from other spectral bands provided in remote sensing datasets. Moreover, existing segmentation methods require large amount of labeled data, which is often scarce in remote sensing.  Several methods have been proposed  to address this issue via pre-training~\cite{neumann2019domain}. Amongst these approaches, self-supervised learning has shown much success in improving the accuracy using limited annotated satellite images~\cite{jean2019tile2vec,vincenzi2020color}.

\subsection{Representation Learning}
Unsupervised learning and self-supervised learning are commonly used to generate feature representation without the need for labour-intensive annotations. Most unsupervised learning methods focus on reconstructing unlabeled data, such as auto-encoders \cite{ranzato2007unsupervised,vincent2008extracting,lee2009convolutional} and deep belief networks (DBN) \cite{le2013building}. In the self-supervised setting, the networks learn discriminative representations after training with pseudo labels created from pretext tasks. The representations learned from such pretext tasks can then be transferred to the downstream tasks. Numerous pretext tasks have been explored in previous literature. For example, image colorization \cite{zhang2016colorful,larsson2017colorization} aims to predict the accurate color version of a photograph, given its gray-scale version as input. Effectively colorizing an image requires the extraction of visual features to capture the semantic understanding of the objects and therefore, visual features can be learned by accomplishing this task. Several deep-learning approaches have been proposed for deep image colorization models~\cite{zhang2016colorful,larsson2017colorization,zhang2017split,larsson2016learning}. Recently this technique has been adopted in the RS domain~\cite{vincenzi2020color}, where an auto-encoder is used to predict RGB channels given the input from other channels. 

Another direction for pretext task, which is commonly used in Natural Language Processing, is the representation learning based on context-similarity~\cite{mikolov2013distributed, pennington2014glove}, where the central the idea is that words that appear in similar contexts should have similar representations. By redefining context as spatial neighborhoods, Tile2Vec \cite{jean2019tile2vec} used this idea in the RS domain where it promotes nearby tiles to have similar representations than the tiles that are far apart. Other popular pretext tasks used in computer vision include image inpainting \cite{pathak2016context}, solving image-jigsaw \cite{noroozi2016unsupervised}, learning by counting \cite{noroozi2017representation}, predicting rotations \cite{gidaris2018unsupervised}, etc. For a comprehensive understanding of Self-supervised representation learning, we would like to redirect the reader to this survey \cite{jing2020self}.

Clustering has also been used used for representation learning. In \cite{yang2016joint}, the authors propose a recurrent framework for clustering and optimises a triplet loss for joint representation learning and clustering. DEC~\cite{xie2016unsupervised} starts with an initial feature representation and cluster assignment, and then iteratively refines both based on the confident samples based on the Kullback-Leibler (KL) divergence loss. One major drawback of these approaches is its tendency to map arbitrary data samples into the same cluster due to the lack of a criteria which respect the local information in image patches. We introduce a reconstruction loss that helps preserve the local information which is essential for semantic-segmentation. 

\section{Problem definition and preliminaries}
\label{Sec:Problem definition}
In this section, we will introduce the available data and our objective. We will also briefly describe the general structure of the segmentation network.
\begin{figure*}
    \centering
    \includegraphics[width=0.8\textwidth]{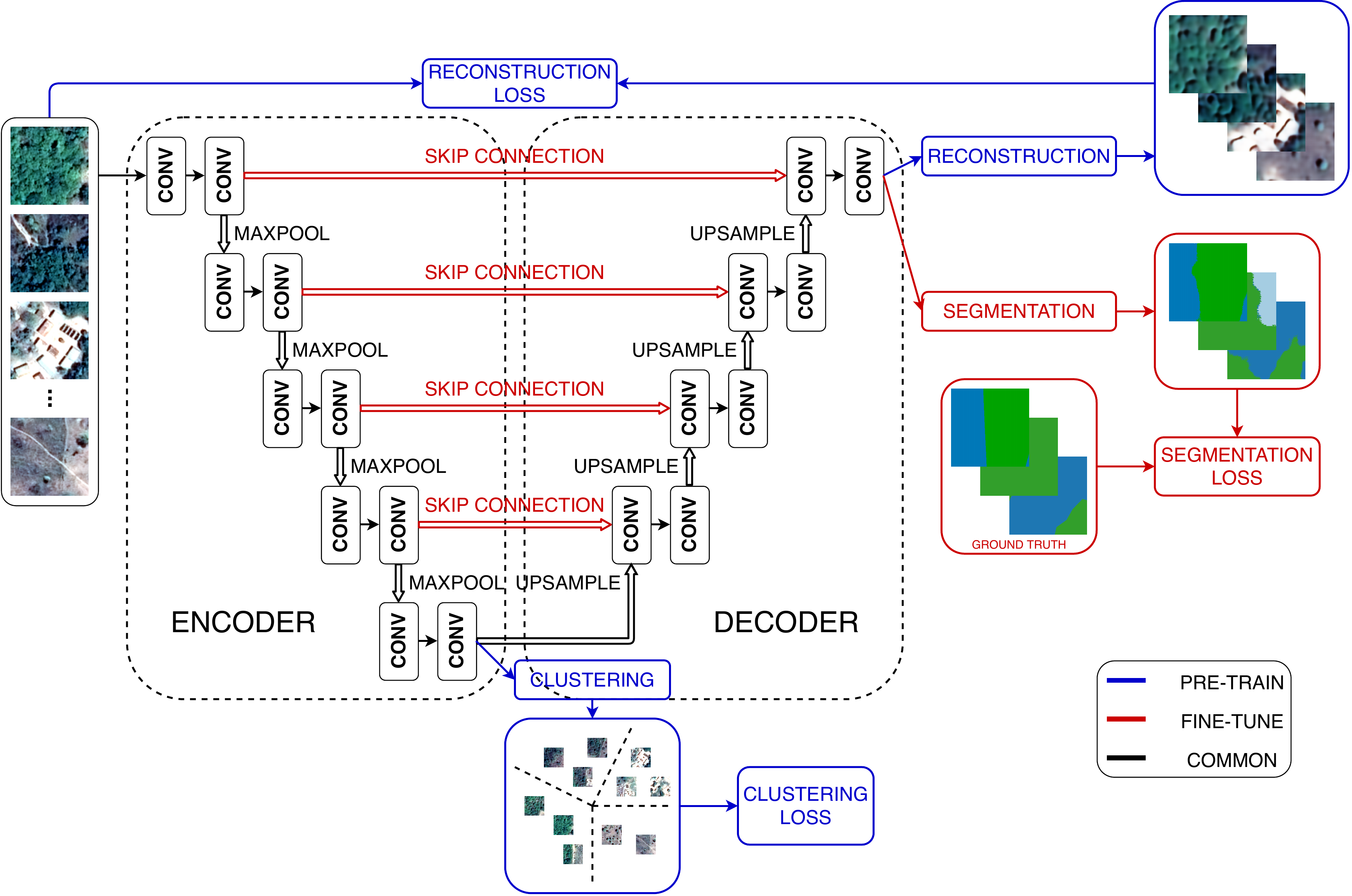}
    \caption{Illustration of the self-supervised pre-trained architecture(best viewed in color). The components that are specifically present during the Pre-training and Fine-tuning stage are drawn in blue and red respectively, while the common components of these two stages are drawn in black. During the self-supervised pre-training step, the skip connections are removed and the classification layer is replaced by a reconstruction layer. These components, highlighted in red, are added back while fine-tuning using the limited labeled samples.}
    \label{fig:architecture}
\end{figure*}

\subsection{Problem setting}
We consider the task of land cover mapping and frame it as a semantic segmentation problem, with the goal of predicting the land cover class of each pixel using the multi-spectral satellite/aerial imagery. In particular, we aim to predict the land cover class $\boldsymbol{l} \in \{1,...,L\}$ of each pixel in an image. During the training process, we have access to limited labeled data and sufficient unlabeled data, which can be described as follows:
\begin{itemize}
    \item[1.] Limited labeled dataset with features and ground truth labels given as $\mathbf{X^{l}} = [X_1^l, \dots, X_{N_l}^l]$ where $X_i^l \in \mathbb{R}^{H\times W\times C}$ is an aerial/satellite image of size $(H,W)$ and having $C$  multi-spectral channels, and $\mathbf{Y^{l}} = [Y^1_l, \dots, Y_{N_l}^l]$ where $Y_i^l \in \mathbb{R}^{H\times W\times L}$ and $L$ is the number of land-cover classes.
    \item[2.] Unlabeled dataset with features given as $\mathbf{X^{u}} = [X_1^u, \dots, X_{N_u}^u]$ where, $X_i^u \in \mathbb{R}^{H\times W\times C}$. Due to the relatively high cost involved in labeling, it is more likely that $N_u >> N_l$. 
\end{itemize}

\subsection{Segmentation network}
A segmentation network $f(X_i;\theta)$ aims to predict the  label of each pixel for an image $X_i$. The parameter $\theta$ is estimated through a training process on 
a fully labeled dataset 
by minimizing an objective function of empirical risk, such as the pixel-wise cross entropy, as follows:
\begin{equation}
    \label{eq:CrossEntropy}
    \mathcal{L}(\boldsymbol{\theta}|\mathbf{X^l},\mathbf{Y^l}) = -\frac{1}{NHW}\sum_i\sum_{(h,w)}\sum_{c} (Y_i)_{h,w}^c\log f(X_i;\theta)_{h,w}^c
\end{equation}
where, $f(X_i;\theta)_{h,w}^c$ is the likelihood of the $(h,w)$’th pixel belonging to class $c$ as predicted by the fully-convolutional network and $(Y_i)_{h,w}^c = 1$ if the $(h,w)$’th pixel of image $i$ belongs to the class $c$.

\section{Method}
\label{Sec:Method}
In this section, we will describe our proposed method CAS. Annotating the multi-spectral images is a labour intensive process and often the labelled dataset do not capture the heterogeneity of the earth due to differences in atmospheric conditions, geography and season when the image was captured. As a result the DNN model learned fail to generalize over the earth's surface. We start with describing the proposed self-supervised learning method CAS using large scale unlabeled data. We then discuss fine-tuning the pre-trained network using limited labeled dataset and the applications in few-shots learning and active learning. 

In this paper, we use the UNet architecture \cite{ronneberger2015u} which consists of an encoder and a decoder, thus, formulating the segmentation function $f(X_i;\theta)$ as a composition of two functions as follows:
\begin{equation}
    f(X_i;\theta) = g(h(X_i; \theta_h);\theta_g)
\end{equation}
where, $h(X_i; \theta_h)$ is the encoder function with  parameters $\theta_h$ which map the input image $X_i$ to an embedding space and, $g(\,\cdot\,; \theta_g)$ is the decoder functions with parameters $\theta_g$ which maps the embeddings back to the image domain.

\subsection{Clustering-Augmented Self-supervised Learning (CAS)}
The UNet model trained from scratch using limited labeled samples can easily overfit the training data. Hence, the learned embeddings become less informative which leads to a poor generalizability of the UNet model. We propose to use a clustering-based pretext learning task to help extract meaningful representation that helps address the land cover heterogeneity. In particular, we adapt DEC as the clustering method, which uses the clustering structure obtained at the image-patch level to naturally separate different land cover modes. We also use additional reconstruction loss to preserve fine-level image details and avoid degenerate solutions (e.g., collapsed clusters) resulting from the standard DEC. Both the DEC and the reconstruction objective are optimized during the self-supervised learning (i.e., model pre-training). In the following, we will describe the details of these involved components.

\subsubsection{Representation Learning with Clustering}
The objective of self-supervised training is to pre-train the segmentation model to extract embeddings that naturally separate image patches with different land cover distributions. In CAS, such representation learning is conducted using large unlabeled dataset in two steps: Phase 1 - model initialization and Phase 2 - representation learning with clustering objective. In the first phase, we use the encoder-decoder from our UNet model and modify it by removing the skip connections and replacing the last classification layer by a reconstruction layer. This modified UNet model is tasked to reconstruct input images. By removing the skip connections, we handicap the use of input information in the reconstruction process, which forces the encoder-decoder model to extract better quality embeddings that fully capture representative features to reconstruct the image without the additional help from the skip connections. In this phase the model is trained by minimizing the following loss function:
\begin{equation}
    \label{eq:Reconstruction}
    \min \frac{1}{N_t}\sum_{i=1}^{N_t} \|g(h(X_i; \theta_h); \theta_g) - X_i\|_2^2,
\end{equation}
where $X_i \in X^l \cup X^u$ and  $N_t = (N_l+N_u)$. Given the obtained embeddings, we conduct KMeans clustering  in the embedding space  by minimizing the following loss function:
\begin{equation}
    \label{eq:Clustering}
    \begin{split}
        \min \frac{1}{N_t}\sum_{i=1}^{N_t} \|g(h(X_i; \theta_h); \theta_g) - Ms_i\|_2^2\\
        \quad s.t.\quad s_{i} \in \{0,1\}^K, 1^Ts_i = 1 \forall i,
    \end{split}
\end{equation}
where ${s_i}$ is the assignment vector for the $i$'th data point, $K$ is the number of clusters, and the $k$'th column of $M$ is the centroid of the $k$'th cluster. The pre-trained autoencoder along with the cluster centroids provide a good initialization point for the encoder parameters $\theta_h$ and cluster centroids $M$.

In the second phase, the encoder parameters and the centroids are refined by learning from the high confidence assignments using an Expectation-Maximisation (EM) style algorithm inspired by the previous work~\cite{xie2016unsupervised}. In the E step the cluster assignment and the target assignment are computed while keeping the encoder parameters and cluster centroids fixed. Specifically, we use a soft-assignment based on the similarity of the embedded data point with the cluster centroid, measured using the Student's t-distribution \cite{maaten2008visualizing}. Specifically, the soft-assignment of data $i$ to cluster $j$ is computed as follows: 
\begin{equation}
    \label{eq:ClusterAssignment}
    q_{ij} = \frac{(1+ \|h(X_i;\theta_h) - M_j\|^2/\alpha)^{\frac{\alpha+1}{2}}}{\sum_{j'=1}^{K}(1+ \|h(X_i;\theta_h) - M_{j'}\|^2/\alpha)^{\frac{\alpha+1}{2}}}
\end{equation}
where $h(X_i;\theta_h)$ is the embedded data point, $\alpha$ is the degree of freedom which is set as 1 in our experiments, and $q_{ij}$ is the probability of assigning the $i$'th data point to the $j$'th cluster. To strengthen prediction and to promote learning from data-points which are assigned with high confidence, the target assignment is computed as:
\begin{equation}
    \label{eq:TargetAssignment}
    p_{ij} = \frac{q^2_{ij}/\sum_i q_{ij}}{\sum_{j'=1}^{K} (q^2_{ij'}/\sum_i q_{ij'})}
\end{equation}

Once cluster assignment and the target assignment are computed, in the M step we estimate the encoder parameters and the cluster centroids  using gradient descent while keeping the cluster and the target assignment fixed. The objective is defined as the KL divergence loss between the soft assignments and the target assignment as follows:
\begin{equation}
    \label{eq:KLLoss}
    \min KL(P\|Q) = \min \frac{1}{N_t}\sum_{i=1}^{N_t}\sum_{j=1}^{K} p_{ij}\log\frac{p_{ij}}{q_{ij}}
\end{equation}

\begin{figure}[h]
    \centering
    \begin{subfigure}[b]{\linewidth}
    \centering
        \includegraphics[width=\linewidth]{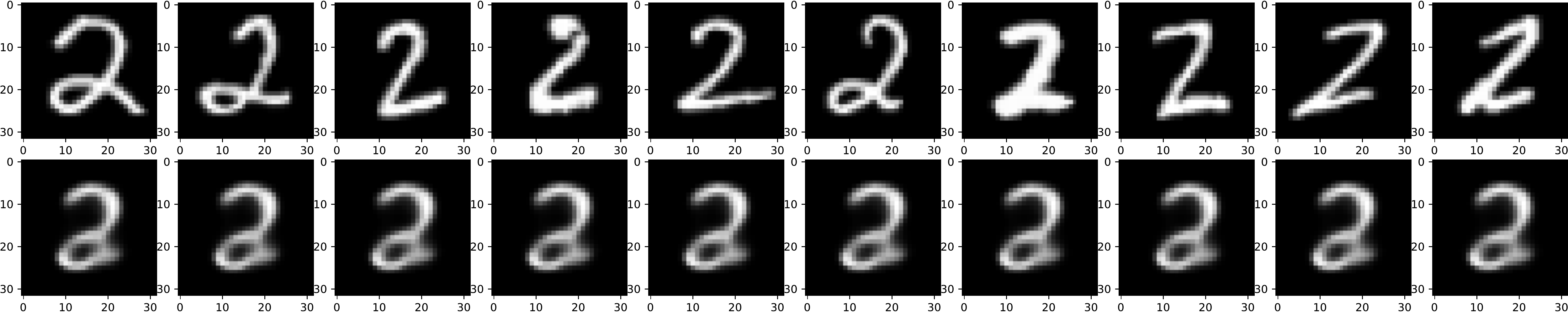}
        \subcaption{Top 10 images from a cluster formed using DEC \cite{xie2016unsupervised}. Top: Input image. Bottom: Reconstructed Image. Note the similar reconstructed images for different input images, which shows the loss of fine-level details.}
    \end{subfigure}
    
    \begin{subfigure}[b]{\linewidth}
    \centering
        \includegraphics[width=\linewidth]{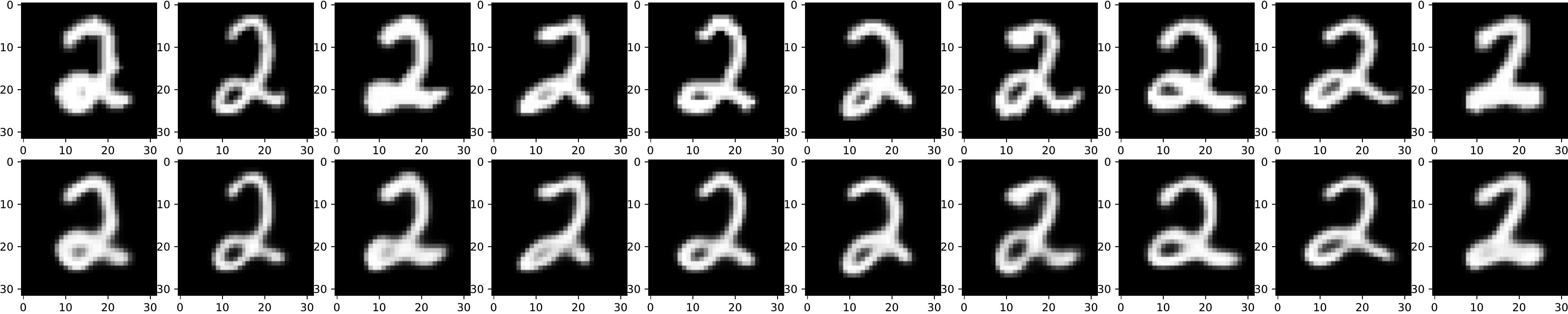}
        \subcaption{Top 10 images from a cluster obtained by CAS. Top: Input image. Bottom: Reconstructed Image. Here we see that the reconstructed images preserve  fine-level details.}
    \end{subfigure}
    \caption{The reconstructed images for the same class using the embeddings learned by (a) DEC and (b) CAS.}
    \label{fig:local details}
\end{figure}

The proposed method faces a number of issues for their use in the semantic-segmentation problem setting. First, there is no provision to avoid degenerate solutions, where the model parameters learned for cluster centroids lead to a trivial solution with the clusters collapsed to a single entity and the representations being zeroed. Second, this approach cannot handle the special scenario where arbitrary data samples are mapped to tight clusters. Finally, since this approach is only to optimize the clustering performance, it forces the embeddings of the data points in the same cluster to be very similar, where we start to lose the finer details of original input images. This is evident from the similar reconstruction of the embedding vectors from two different images from the same class as shown in figure \ref{fig:local details} (a). This loss of fine-level image details becomes a serious issue in the semantic segmentation problem since we aim to assign a label to each pixel in the image instead of assigning a single label to the entire image as in the image classification setting.

\subsubsection{Preserving fine-level details}
To enable learning from the confident samples while also preserving the finer details and overcome the issues mentioned in the previous subsection, CAS augments the KL Divergence based clustering loss with the reconstruction loss. Specifically, we add a decoder that reconstructs the data-point using the embeddings while the clustering task is performed at the bottle-neck layer. The encoder parameters, decoder parameters and the cluster centroids are refined according to the objective:
\begin{equation}
    \mathcal{L} = \frac{1}{N_t}\sum_{i=1}^{N_t}\left(\sum_{j=1}^{K} p_{ij}\log\frac{p_{ij}}{q_{ij}} + \lambda \|g(h(X_i; \theta_h); \theta_g) - X_i\|_2^2\right),
\end{equation}
where $\lambda$ is a hyper-parameter to balance the clustering loss and the reconstruction loss. 

The proposed modifications provides a number of benefits. First, reconstruction loss prevents the model to collapse to a degenerate solution by ensuring that the decoder can reconstruct the data point using the embeddings. Second, since the decoder has to reconstruct the images from the embeddings, it prevents the embeddings to lose the fine-level details thus helping in the segmentation. Finally, the trained decoder provides as a good initialization for the decoder of the segmentation network.

\subsection{Downstream applications}
After obtaining the pre-trained model through self-supervised learning, we describe two downstream applications where we use labeled data to fine-tune the model. 

\subsubsection{Few-shots segmentation}
After training the encoder-decoder model, we feed the learned weight parameters to the U-Net segmentation model with skip connections (see Fig.~\ref{fig:architecture}). This model can be fine-tuned using pixel-wise labels by minimizing the cross-entropy loss using labeled data (see Eq.~\ref{eq:CrossEntropy}).

\subsubsection{Active learning}
The clustering structure extracted by the proposed method also enables actively select query image patches so as to reduce the manual efforts in data labeling. The objective is to select a small number of query image patches to ask for labeling so that the performance of segmentation model is optimized after it is trained with these labeled patches. In particular, we uniformly select image patches from different clusters that are closest to cluster centroids.  Since the clustering structure automatically divides the whole data space into $K$ disjoint set of data points, uniformly selected patches are representative samples that cover different types of data in the entire data space. 

Furthermore, we can extend this approach to handle the scenario where the budget (i.e., the number of query samples) is not divisible by the number of clusters. In this case, we aim to take more samples from clusters of higher uncertainty.  Intuitively, each cluster contains images with similar data distribution and thus the labels predicted by a well-trained segmentation model should be similar for all the images within a cluster. Specifically, we first predict pixel-wise labels for all the images and then  estimate the majority class for each image. We measure the uncertainty of each cluster $k$ as the entropy of these obtained majority classes. 

\begin{table*}[ht]
    \centering
    \caption{Comparison with baselines in terms of Mean F1 Score (and standard deviation) with increasing number of samples. The last row (All Data) shows the performance of using all the available data for supervised training (without pre-training).}
    \label{Tab:tableResults}
    \begin{tabular}{|>{\centering\arraybackslash}p{1.7cm}||>{\centering\arraybackslash}p{0.7cm}|>{\centering\arraybackslash}p{0.7cm}|>{\centering\arraybackslash}p{0.7cm}|>{\centering\arraybackslash}p{0.7cm}|>{\centering\arraybackslash}p{0.7cm}|>{\centering\arraybackslash}p{0.7cm}||>{\centering\arraybackslash}p{0.7cm}|>{\centering\arraybackslash}p{0.7cm}|>{\centering\arraybackslash}p{0.7cm}|>{\centering\arraybackslash}p{0.7cm}|>{\centering\arraybackslash}p{0.7cm}|>{\centering\arraybackslash}p{0.7cm}|}
        \hline
        \textbf{} & \multicolumn{6}{c||}{\textbf{D1: Cashew Plantation Mapping}} & \multicolumn{6}{c|}{\textbf{D2: Crop Mapping}} \\\hline
        \textbf{Method} & 10 & 20 & 40 & 120 & 160 & 200 & 10 & 20 & 50 & 100 & 150 & 200\\\hline\hline
        OnlyLabeled & $\underset{(0.098)}{0.402}$ & $\underset{(0.059)}{0.572}$ & $\underset{(0.050)}{0.609}$ & $\underset{(0.021)}{0.704}$ & $\underset{(0.018)}{0.712}$ & $\underset{(0.017)}{0.724}$ & $\underset{(0.121)}{0.426}$ & $\underset{(0.073)}{0.634}$ & $\underset{(0.047)}{0.700}$ & $\underset{(0.016)}{0.788}$ & $\underset{(0.015)}{0.809}$ & $\underset{(0.014)}{0.837}$\\ \hline
        AutoEncoder & $\underset{(0.098)}{0.481}$ & $\underset{(0.053)}{0.629}$ & $\underset{(0.035)}{0.663}$ & $\underset{(0.026)}{0.717}$ & $\underset{(0.018)}{0.737}$ & $\underset{(0.016)}{0.743}$ & $\underset{(0.139)}{0.508}$ & $\underset{(0.054)}{0.666}$ & $\underset{(0.051)}{0.722}$ & $\underset{(0.016)}{0.798}$ & $\underset{(0.013)}{0.814}$ & $\underset{(0.007)}{0.839}$\\ \hline
        Tile2Vec & $\underset{(0.048)}{0.507}$ & $\underset{(0.021)}{0.632}$ & $\underset{(0.024)}{0.686}$ & $\underset{(0.008)}{0.739}$ & $\underset{(0.008)}{0.740}$ & $\underset{(0.008)}{0.745}$ & $\underset{(0.057)}{0.566}$ & $\underset{(0.026)}{0.688}$ & $\underset{(0.026)}{0.757}$ & $\underset{(0.017)}{0.800}$ & $\underset{(0.014)}{0.825}$ & $\underset{(0.004)}{0.841}$\\ \hline
        Colorization & $\underset{(0.044)}{0.609}$ & $\underset{(0.037)}{0.660}$ & $\underset{(0.013)}{0.710}$ & $\underset{(0.008)}{0.756}$ & $\underset{(0.004)}{0.762}$ & $\underset{(0.004)}{0.776}$ & $\underset{(0.055)}{0.543}$ & $\underset{(0.046)}{0.678}$ & $\underset{(0.039)}{0.729}$ & $\underset{(0.014)}{0.789}$ & $\underset{(0.011)}{0.823}$ & $\underset{(0.007)}{0.837}$\\ \hline
        DEC & $\underset{(0.024)}{0.628}$ & $\underset{(0.016)}{0.688}$ & $\underset{(0.016)}{0.709}$ & $\underset{(0.008)}{0.747}$ & $\underset{(0.008)}{0.751}$ & $\underset{(0.007)}{0.756}$ & $\underset{(0.043)}{0.600}$ & $\underset{(0.023)}{0.723}$ & $\underset{(0.019)}{0.763}$ & $\underset{(0.008)}{0.814}$ & $\underset{(0.007)}{0.837}$ & $\underset{(0.007)}{0.843}$\\ \hline
        CAS(ours) & $\underset{(0.030)}{\textbf{0.674}}$ & $\underset{(0.020)}{\textbf{0.721}}$ & $\underset{(0.008)}{\textbf{0.736}}$ & $\underset{(0.007)}{\textbf{0.767}}$ & $\underset{(0.008)}{\textbf{0.774}}$ & $\underset{(0.002)}{\textbf{0.783}}$ & $\underset{(0.058)}{\textbf{0.656}}$ & $\underset{(0.024)}{\textbf{0.759}}$ & $\underset{(0.010)}{\textbf{0.792}}$ & $\underset{(0.007)}{\textbf{0.831}}$ & $\underset{(0.004)}{\textbf{0.845}}$ & $\underset{(0.002)}{\textbf{0.847}}$\\ \hline
        All Data & \multicolumn{6}{c||}{0.795 (1500 patches)} & \multicolumn{6}{c|}{0.87 (700 patches)} \\\hline
    \end{tabular}
    \vspace{-0.2in}
\end{table*}

\section{Experimental Results}
\label{Sec:Experimental Results}

We evaluate our proposed strategy for semantic segmentation on two real-world applications of great societal impacts. In the first example, we aim to map cashew plantation in Benin, which contribute nearly 10\% of the country's export income. Benin government is actively looking for inventory information of cashew to assist the distribution of their recent \$100 million loan from World Bank, aiming at further developing the cashew industry. In the second example, we investigate crop mapping in the US Midwest, the world's bread basket. Mapping crops is a key step towards many applications, such as forecasting yield, guiding sustainable management practices and evaluating progress in conservation efforts.

\subsection{Datasets}
\begin{itemize}
    \item[D1:] \textbf{Cashew Plantation Mapping}
    We use the multi-spectral images captured by AIRBUS in 2018 to study an area in Africa. The images have 4 spectral bands namely red, green, blue and NIR (near infrared) at a spatial resolution of 0.5 metres. For our experiment, we divide our study region into patches of size $68\times 68$ and each pixel within this patch is assigned a class label $l \in \{$ Cashew, Forest, Urban, Background $\}$. The  ground  truth  was  created  using manual annotation over the entire study region provided by our collaborators in Benin, Africa~\footnote{Given the proprietary nature of the Planet Lab composite and the Airbus imagery, we do not have permission to make this data publicly available.}.
    
    \item[D2:] \textbf{Crop Mapping}
    We used publicly available multi-spectral images observed by the Sentinel-2 Constellation. The Sentinel-2 data product has 13 spectral bands~\footnote{\url{https://developers.google.com/earth-engine/datasets/catalog/COPERNICUS_S2_SR##bands}} at three different spatial resolutions of 10, 20 and 60 metres. For consistency, bands with 20 and 60 metres resolution are resampled by using the nearest neighbour method to 10 metres. For our experiment, we consider the region of southwestern Minnesota,US, where we aim to classify each pixel to a class label $l \in$ \{ Corn, Soybean, Sugarbeats, Water, Urban \}. Our data is taken in August 8, 2019.  The labels are obtained from the USDA Crop Data Layer product~\cite{cdl}.
\end{itemize}

\subsection{Baselines}
\label{sec:Baseline}
We use the UNet architecture as the base model for semantic segmentation and compare our representation learning strategy against the following baselines. Here all the representation learning methods are trained on the entire training set (labeled + unlabeled data).  
\begin{enumerate}
    \item \textbf{OnlyLabeled} This method considers training a UNet from scratch, only using the labeled dataset.
    \item \textbf{AutoEncoder} We pre-train the UNet model by transforming it into an autoencoder structure by removing the skip connections and conduct reconstruction in the final layer (described in Section \ref{Sec:Method}).
    \item \textbf{Tile2Vec} We adopt this method~\cite{jean2019tile2vec} to learn representation by leveraging spatial contextual similarities. To prevent the model from collapsing and providing degenerate solution, we initialize the model using the AutoEncoder baseline. The model is optimized using a triplet loss among the achor, neighbors and distant patches.
    \item \textbf{Colorization}~\cite{vincenzi2020color} The segmentation model has two independent branches which takes in the spectral bands and the RGB channels, respectively. The first branch is pre-trained using the colorization task and the second branch is pre-trained on ImageNet \cite{deng2009imagenet}. As proposed by the authors, both of the branches are fine-tuned separately on the limited labeled samples and we average their predictions as final outputs.
    \item \textbf{DEC} 
    We adopt the method presented in~\cite{xie2016unsupervised} to learn representations that optimizes a clustering-based loss. This optimisation is performed at the image patch-level and thus disregards the fine-level image details.
\end{enumerate}

\begin{figure*}
    \centering
    \includegraphics[width=0.8\textwidth, height=10cm]{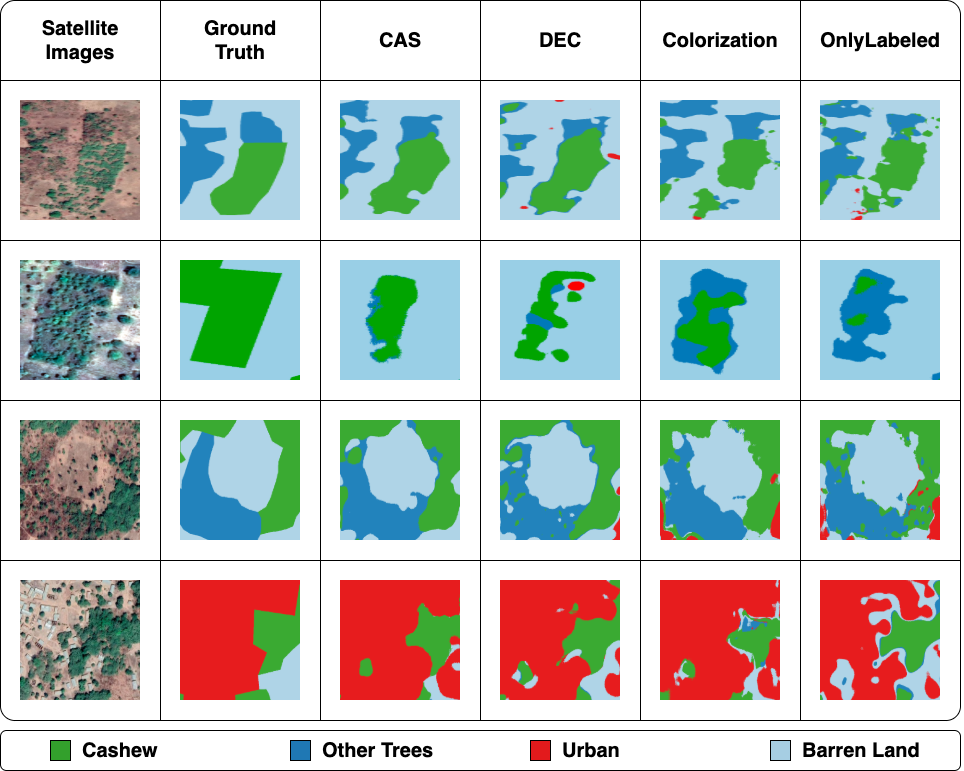}
    \vspace{-0.1in}
    \caption{Examples of land cover mapping made by different methods. The first column shows the reference RGB images and the second column shows the manually-created ground-truth data. }
    \label{fig:Segmentation Maps}
    \vspace{-0.2in}
\end{figure*}

\subsection{Few-Shot Learning}
Here we evaluate the methods for few-shot learning setting where we progressively increase the number of labelled samples for training. The average accuracy and standard deviation of 5 runs for all the algorithms are reported in Table \ref{Tab:tableResults}. The model trained from scratch using only labelled instances (\textit{OnlyLabeled}) performs the worst. \textit{AutoEncoder} takes advantage of the larger unlabelled dataset in learning the representations and thus shows an increase in performance than \textit{OnlyLabeled}. The representations learned by \textit{OnlyLabeled} and \textit{AutoEncoder} do not capture discriminative information of land covers and thus they do not perform as well as \textit{DEC}. The next baselines of \textit{Tile2Vec} and \textit{Colorization} makes use of alternate ways of representation learning on the unlabelled data as described in section \ref{sec:Baseline}. Each of these provide limited improvement over AutoEncoder. \textit{Tile2Vec} uses assumptions that the nearby spatial tiles are similar and far away are different which can sometimes be inaccurate. \textit{Colorization} learns representations by learning to colorize the images which can sometimes be ineffective in distinguishing the regions where the color is not distinctive. Next, we see that our adaptation of \textit{DEC} (that captures information about different types of land covers via clustering) is able to do nearly as well or better (especially for small number of samples) than the schemes such as \textit{Colorization} that are able to explicitly preserve fine levels details. Finally, our proposed scheme \textit{CAS} outperforms all these baselines. 

In Fig.~\ref{fig:Segmentation Maps}, we show the mapping results of different methods in several example regions from D1. The segmentation results shown are obtained from the models trained using 40 labeled samples. We can see the detection results produced by CAS are more consistent to the ground truth and the satellite images. In contrast, other self-supervised learning  methods (DEC and Colorization) often cannot precisely delineate land cover boundaries. This is because the plantations that are close the boundary commonly have lower density and thus are more likely to be confused with other land covers. 

\subsubsection{Effect of more labeled training samples:}
Due to the limited number of labeled samples in the downstream task, the performance of the models trained from scratch depend on the representability of those small subset of data points. The limited data samples do not capture the whole data domain and thus the representations learned using them are not robust. Self-supervised learning aims to decouple the representation learning phase and the classification phase. \textit{CAS} tries to leverage the unlabeled data to capture the representations and then learn the classification rules using the limited dataset. With the increase in the number of labeled instances, the representations learned using them become increasingly more robust. This results in a reduction in the gain obtained by using the unlabeled data in the representation learning manner. This is evident from the result shown in Table \ref{Tab:tableResults}, where we increase the number of labeled patches for both the datasets. We observe that the accuracy of all methods increase with the increase in the number of labeled patches.

\begin{figure}
    \centering
    \begin{subfigure}[b]{0.48\linewidth}
        \includegraphics[width=\linewidth]{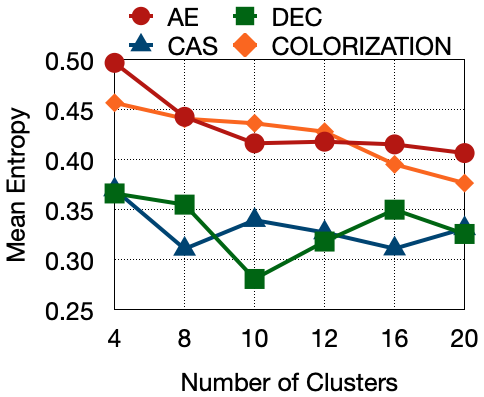}
        \subcaption{}
    \end{subfigure}
    \begin{subfigure}[b]{0.48\linewidth}
        \includegraphics[width=\linewidth]{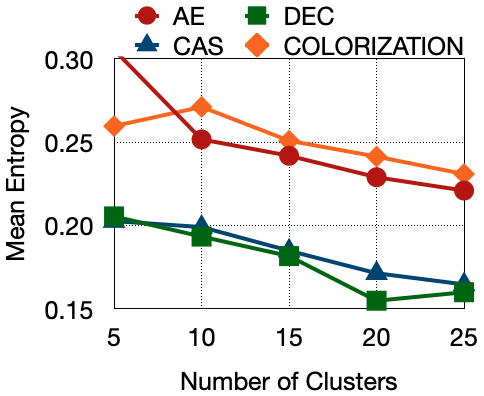}
        \subcaption{}
    \end{subfigure}
    \caption{Average entropy of the clusters obtained by different methods on Dataset (a) D1 and (b) D2.}
    \label{fig:entropy}
    \vspace{-0.2in}
\end{figure}

\begin{figure}[!t]
    \centering
    \includegraphics[width=\linewidth]{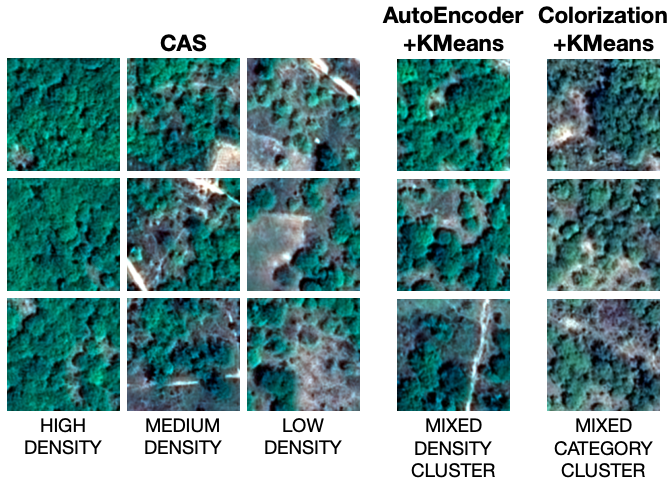}
    \vspace{-0.3in}
    \caption{First three columns are seperate clusters formed by CAS which clearly show a clusters of high, medium and low density. The last two columns are one of the clusters formed by AutoEncoder and Colorization respectively.}
    \label{fig:clusters}
    \vspace{-0.2in}
\end{figure}

\subsection{Clustering-based Evaluation of Representations}
Here we evaluate the quality of representation produced by different approaches using the quality of clustering produced using them. Specifically, we measure the clustering performance using aggregated labels of image patches. For each image patch, we define the aggregated label as the majority label from all the pixels of this image patch. Intuitively, we expect image patches within a cluster to have have the same aggregated labels. Hence, we estimate the clustering performance using the weighted entropy of aggregated labels. Specifically, given a clustering structure, we first compute the entropy of aggregated labels for each cluster. Then we compute the weighted average of entropy values over all the clusters based on their cluster sizes. The lower value of the average entropy indicates better clustering performance. 

We compare the clusters extracted by the baselines with our proposed method (Fig.~\ref{fig:entropy}). In \textit{AutoEncoder} and \textit{Colorization}, KMeans clustering is conducted on the obtained embeddings. It can be seen that the proposed method significantly outperforms \textit{Autoencoder} and \textit{Colorization} in both datasets. \textit{DEC} and our proposed method \textit{CAS} achieve very pure clusters even using no more than five clusters. Besides, our method achieves similar performance with \textit{DEC} even though we simultaneously optimize the clustering performance and the reconstruction error. Although, \textit{DEC} achieves good clusters, it is plagued with the issues highlighted in Fig.~\ref{fig:local details}, which hampers its segmentation performance.

An example of the clusters formed by the methods are shown in Fig.~\ref{fig:clusters}. We observe that the clusters formed by \textit{CAS} capture the intra-class heterogenity and form pure clusters, while the other cluster formed by the other methods highlight several issues which we motivated in the introduction. As shown in Fig.~\ref{fig:clusters}, one of the clusters formed by \textit{AutoEncoder} has a mixture of high, medium and low density clusters which points towards the intra-class confusion. The images of the cluster formed by \textit{Colorization} are covered by other trees, low-density cashew and mixture of other trees and cashew respectively.  This highlights the inter-class confusion due to plantations being confused with other trees.

\begin{figure}
    \centering
    \begin{subfigure}[b]{0.49\linewidth}
        \includegraphics[width=\linewidth]{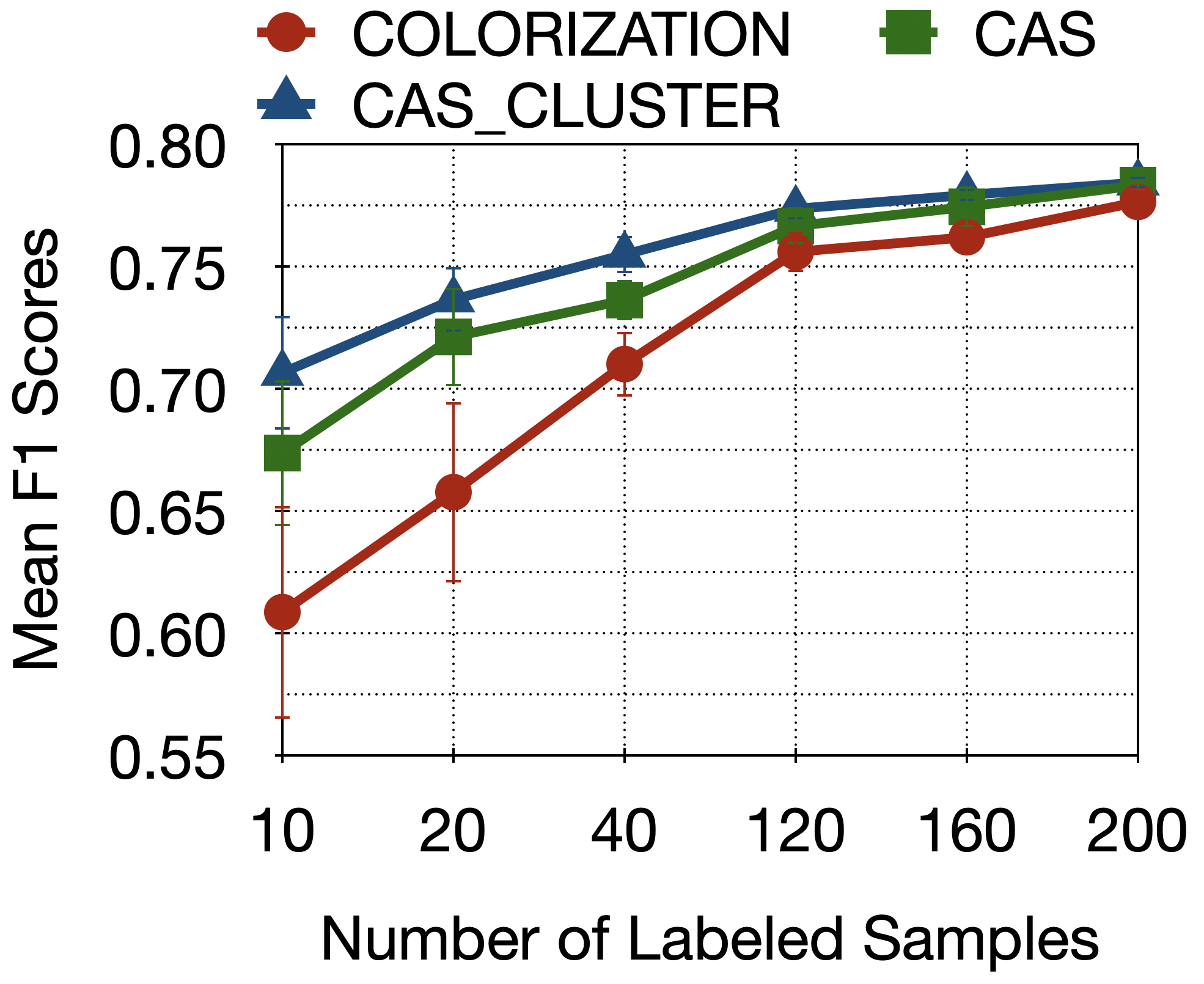}
        \subcaption{}
    \end{subfigure}
    \begin{subfigure}[b]{0.49\linewidth}
        \includegraphics[width=\linewidth]{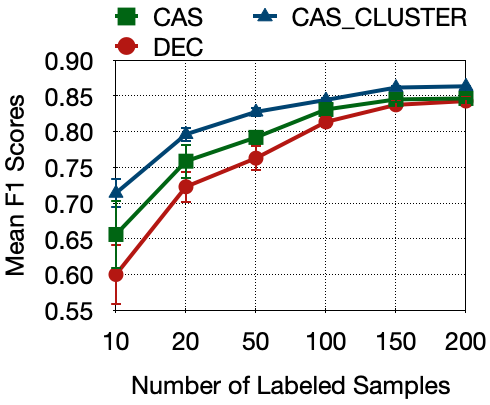}
        \subcaption{}
    \end{subfigure}
    \vspace{-.1in}
    \caption{Our method is compared with the next best method while using active learning on Dataset (a) D1 and (b) D2. CAS\_CLUSTER represents the method to actively sample from clusters obtained from CAS.}
    \label{fig:activeSampling}
\end{figure}

\subsection{Using Clusters for Active Sampling}
Here we show the effectiveness of the active learning strategy.  In particular, we use obtained clusters to query patches rather than randomly sampling patches for labeling.  Fig.~\ref{fig:activeSampling} shows the segmentation performance when we label different amount of samples either using our active learning approach or by using random sampling. We also show the performance of random sampling both for \textit{CAS} model and the best-performing baseline in each dataset (\textit{Colorization} in D1 and \textit{DEC} in D2). 

According to the segmentation performance, we can observe that the active learning method leads to better performance, especially when we only label small amount of samples. This demonstrate the effectiveness of using the clustering structure obtained from CAS to select most representative samples given a limited budget. When we label sufficient amount of samples ($>$200 samples), all the methods achieve similar performance. 

\section{Conclusion}\label{Sec:Conclusion}
In this paper we propose the use of clustering based self-supervised learning to pre-train the model for few-shot segmentation. This method is able to preserve fine-level details while also  extracting a clustering structure to naturally separate heterogeneous land cover modes. The obtained clustering structure can also be used in an active learning setting. We conduct experiments on two real world datasets related to land-cover mapping to show the benefits brought by using the abundant unlabeled data. Further, we compare our method with other forms of self-supervised learning strategies adopted in the Remote Sensing domain, namely Colorization and Tile2Vec, to show the effectiveness of our proposed strategy. 

Given the effectiveness of our proposed method in mapping heterogeneous land covers using limited labels, our framework has the potential for creating large-scale (e.g., global) land-cover maps using satellite imagery and small amount of manually-created labels.  
Moreover, our proposed framework can be generally applied to a variety of spatial datasets (e.g., traffic and crime data) which exhibits strong heterogeneity.  

Although our proposed method has produced improved accuracy in land cover mapping, it remains limited in discovering temporal patterns from multi-temporal satellite data which is often available in public satellite datasets. Another important direction is to combine the pretext task of clustering with pretext tasks that is defined to reflect land cover distinctions based on domain knowledge.

\section{Acknowledgements}
This work was funded by the NSF awards 1838159 and 1739191 and National Aeronautics and Space Administration (NASA) Land Cover Land Use Change program, grant number 80NSSC20K1485. Rahul Ghosh is supported by UMII MNDrive Graduate Fellowship. Access to computing facilities was provided by the Minnesota Supercomputing Institute.

\bibliographystyle{ACM-Reference-Format}
\bibliography{main}

\end{document}